\documentclass[letterpaper]{article}

\usepackage{graphicx,setspace}
\usepackage{hyperref}
\usepackage[margin=1in]{geometry}

\usepackage[numbers,sort&compress]{natbib}

\title{Cardinality Leap for Open-Ended Evolution:\\
Theoretical Consideration and Demonstration by ``Hash Chemistry''}
\author{Hiroki Sayama$^{1,2}$\\
\mbox{}\\
$^1$ Center for Collective Dynamics of Complex Systems\\
Binghamton University, State University of New York\\
Binghamton, NY 13902-6000, USA\\
$^2$ Waseda Innovation Lab\\
Waseda University, Shinjuku, Tokyo 169-8050, Japan\\
sayama@binghamton.edu}

\begin{document}
\maketitle

\begin{abstract}
Open-ended evolution requires unbounded possibilities that evolving entities can explore. The cardinality of a set of those possibilities thus has a significant implication for the open-endedness of evolution. We propose that facilitating formation of higher-order entities is a generalizable, effective way to cause a ``cardinality leap'' in the set of possibilities that promotes open-endedness. We demonstrate this idea with a simple, proof-of-concept toy model called ``Hash Chemistry'' that uses a hash function as a fitness evaluator of evolving entities of any size/order. Simulation results showed that the cumulative number of unique replicating entities that appeared in evolution increased almost linearly along time without an apparent bound, demonstrating the effectiveness of the proposed cardinality leap. It was also observed that the number of individual entities involved in a single replication event gradually increased over time, indicating evolutionary appearance of higher-order entities. Moreover, these behaviors were not observed in control experiments in which fitness evaluators were replaced by random number generators. This strongly suggests that the dynamics observed in Hash Chemistry were indeed evolutionary behaviors driven by selection and adaptation taking place at multiple scales.\\
Keywords: Open-ended evolution, set of possibilities, cardinality leap, higher-order entities, Hash Chemistry, universal fitness evaluator
\end{abstract}

\section{Introduction}

Open-ended evolution (OEE) \citep{taylor2016open,banzhaf2016defining,taylor2018routes} requires unbounded possibilities that evolving entities can explore. Such an infinite possibility space can be conceptualized mathematically as an infinite set of all possible types, on which the {\em landscape} of evolution is constructed. One can picture that evolving entities collectively search this landscape, over an indefinitely period of time, for locations (possibilities) where they can have a greater chance of continuous existence. It remains unclear, however, how one can effectively achieve such an infinite possibility space in artificial life (ALife) models, and how its cardinality, or the ``size'' of the set of possibilities, affects evolutionary dynamics.

In this paper, we revisit an important fact that the cardinality of possibilities has a significant implication for the open-endedness of evolution, and propose that facilitating formation of higher-order entities is a generalizable, effective way to cause a ``cardinality leap'' in the set of possibilities. We first provide a brief theoretical consideration based on mathematical concepts of cardinalities, and then demonstrate the idea using a simple, proof-of-concept toy model that we call ``Hash Chemistry.''

\section{Theoretical Consideration}

\subsection{Mathematical assumptions}

Let $S$ be the set of all possibilities of individual entities in an evolutionary system, such as all possible genotypes or evolving computer codes. $S$ can be either finite or infinite. For evolution to be open-ended, $S$ should be infinitely large, but in all practical implementations of living systems (including biological and artificial ones) $S$ is technically finite because evolving entities cannot be bigger than the environment. In addition, if the entities are symbolically represented (as is the case in real biological evolution and in most of ALife models), $S$ is made of discrete entities, and therefore it is countable.

Here we assume a very simplistic view of evolution that there is(are) an {\em effectively optimal} possibility(-ies) somewhere in $S$, whose ``fitness'', i.e., chance of successful survival, cannot be surpassed virtually by any other entity. 
We do not call this just {\em optimal} because, theoretically, one can construct a landscape on which no optimal type exists, e.g., when the fitness is given by $1 - 1/L$ where $L$ is the length of the entity's description. But even in such cases, the fitness resolution limit coming from limitations in the laws of the world (e.g., quantum limit) makes two types indistinguishable if their $L$s are very large. Those indistinguishable types are both {\em effectively optimal}.

Evolution can be visualized as a collective search process in $S$ to find such  effectively optimal entities. Once the evolution finds one, that entity will dominate the system, and the open-endedness of evolution will no longer be exhibited because there will be no more adaptive types discovered by the evolutionary process. The question of our interest, then, is if an evolutionary system is destined to reach this final state, and if so, if there is any workaround available to help the system avoid reaching this destination so that it can remain open-ended, especially from a viewpoint of constructing {\em artificial} evolutionary systems.

\subsection{Cardinality and open-endedness}

Assuming that $S$ is countable, all the possibilities can be mapped onto a set of natural numbers (1, 2, 3, ...) in some arbitrary order. The effectively optimal entity is also located somewhere in this set of numbers. Let $n_o$ be the number that corresponds to that entity. If $S$ is finite, it is obvious that there is no way evolution could produce an indefinitely long sequence of search history continuously producing novel types without hitting $n_o$, and therefore, the evolution cannot be open-ended. If $S$ is countably infinite, however, it is mathematically possible for evolution to go on indefinitely without hitting $n_o$ (especially if it carefully avoids $n_o$---although somewhat contrived). Moreover, if $S$ is uncountably infinite, any amount of enumerations of entities in $S$ (which is essentially what evolution does) can have almost zero probability to hit $n_o$ because any enumerations, even continued indefinitely, are only countably infinite at most. These arguments illustrate that the cardinality difference of $S$ has significant implications for the possibility to achieve OEE.

How can one utilize the above mathematical arguments for designing potentially open-ended artificial evolutionary systems? One straightforward approach is simply to adopt real-valued representations of evolving entities. Since enumeration of such continuous values is not possible, true OEE is logically possible within this setting. This view is, however, not consistent with the mechanisms of biological evolution as we know it where entities are encoded in discrete symbols. It also has a risk to fall into a naive conclusion that any chaotic dynamical systems are OEE systems. While it may be true mathematically, such a simplistic conclusion may not be helpful in making advance in ALife research on OEE.

\subsection{Higher-order entities and cardinality leap}

Here we take a different approach to the above issue, which, we believe, is generalizable and more useful for research on OEE. Specifically, we consider a way to expand the cardinality of the possibility set by facilitating formation of {\em higher-order entities}, i.e., combinations/coalitions of multiple individual entities in $S$. 

Mathematically, a higher-order entity can be defined as a multiset (a set in which multiple copies of an identical entry are allowed) of entities of $S$. For example, with $S$ being a set of chemical elements, molecules like $O_2$ and $H_2O$ are higher-order entities, represented by multisets $\{O, O\}$ and $\{H, H, O\}$, respectively. Other examples include a multiset of molecules contained within a micelle, a multiset of organelles contained within a eukaryotic cell, symbiosis of multiple organisms, etc. It can be argued that many, if not all, of the major transitions in evolution \citep{smith1997major,szathmary2015toward} can be described mathematically as formation of such higher-order entities (i.e., multisets of individual entities).

An important mathematical fact we want to point out is that the formation of higher-order entities naturally causes a ``cardinality leap'' in the possibility set. Let $S^*$ be the possibility set of higher-order entities, i.e., the set of all multisets of individual entities in $S$. Each higher-order entity in $S^*$ can be mapped onto a sequence of non-negative integers whose components represent numbers of individual entities of $S$ that are contained in the multiset. 
For example, with $S=\{H, He, Li, Be, B, C, N, O, F, Ne, \ldots \}$, $O_2$ and $H_2O$ are mapped onto $(0, 0, 0, 0, 0, 0, 0, 2, 0, 0, \ldots)$ and $(2, 0, 0, 0, 0, 0, 0, 1, 0, 0, \ldots)$, respectively. Given that the numbers of individual entities that appear in the multiset are theoretically unbounded, the size of $S^*$ is infinite. If $S$ is finite, $S^*$ is the set of $|S|$-dimensional non-negative integer-valued vectors, and thus $S^*$ is countably infinite. Or, if $S$ is already countably infinite, $S^*$ is the set of infinitely long sequences of non-negative integers, whose cardinality equals the cardinality of real numbers (i.e., uncountably infinite) by the diagonal argument \citep{simmons1993universality}.

Note that in either case, the cardinality of possibility sets makes a fundamental leap (from finite to countably infinite, or from countably infinite to uncountably infinite). Such a cardinality leap would greatly promote the open-endedness of evolutionary processes. In particular, if the original possibility set $S$ is already countably infinite, the formation of higher-order entities can produce an uncountably infinite possibility space, achieving OEE very naturally in it, even if the individual entities in $S$ are represented in discrete symbols.

\section{Proof of Concept: Hash Chemistry}

The theoretical consideration given in the previous section may give us some hope, but it does not provide concrete guidelines about how one can construct an artificial evolutionary system that has the potential to form higher-order entities and thereby exhibit OEE. In this section, we present a simple toy model to demonstrate how it could be done, at least in a very primitive form.

\subsection{General architecture}

In creating artificial evolutionary systems that can facilitate formation of higher-order entities, one technical challenge is how to design a mechanistic, universal means that can evaluate the level of success of entities {\em of any arbitrary size}. The real laws of physics/chemistry/biology apparently have no issue on this, because they are fundamentally bottom-up and fully distributed \citep{stepney2005journeys}. However, most of the existing evolutionary models assume a typical genotype-phenotype mapping, evaluating the fitness of each individual type separately \citep{mitchell1994genetic,ofria2004avida,banzhaf2006guidelines,taylor2016open}. Some evolutionary models adopted more elaborated methods such as context/environment dependence, multilevel selection and multiscale interactions \citep{traulsen2006evolution,goodnight2008evolution,fernandez2012emergent}, but the fitness evaluation mechanisms in those models are not flexible enough to be applied to possibility sets like $S^*$ in which the size/order of entities is not bounded. While nature does such universal fitness evaluation so easily, constructing a physically/chemically/biologically plausible artificial mechanism for universal fitness evaluation of any higher-order entities would probably require a combinatorially large amount of design effort.

Because solving the above challenge is not part of our objectives, we circumvent the problem by throwing it at {\em Deus ex Machina} available in most computational environments, a.k.a. the {\em hash function}. A hash function takes any hashable data and returns a hash value that is deterministically assigned to the given data, which is perfect as a quick-and-easy substitute of the universal fitness evaluator\footnote{One might argue that a LISP interpreter also has a similar property suitable for a universal fitness evaluator. However, such universal program interpreters may take very long time to produce a return value (and they may not even halt for some inputs). In contrast, hash functions always produce a return value, and they are fast in general.}. Note that using a hash function for fitness evaluation does not mean fitness values are randomly assigned each time fitness is evaluated. Instead, the hash function always returns the same, consistent value for the same entity type. This consistency provides room for potential adaptation of evolving entities.

With the power of this hash function, one can build a simple evolutionary model, which we call ``Hash Chemistry,'' in the following general architecture that follows a typical Artificial Chemistry framework \citep{dittrich2001artificial,banzhaf2015artificial}:
\begin{enumerate}
\item Define a set of possibilities of individual entities ($S$).
\item Define a spatial domain in which entities reside and interact. This domain can be a continuous Euclidean space, a discrete lattice or network, or any other space, as long as one can represent proximity of entities on it.
\item Initialize the system by placing some individual entities in the spatial domain.
\item Using any method of choice, extract a multiset of individual entities that are spatially close to each other (this could be just a set of one entity as well). This multiset is now a higher-order entity whose behavior is to be determined below.
\item Apply the hash function to the multiset extracted above (typically represented as a sorted list of the types of individual entities in it). The returned hash value should be normalized somehow to an interpretable fitness value $f$.
\item Based on $f$, do one of the following:
(a) Add a copy of the multiset to the space (replication).
(b) Remove the extracted multiset from the space (death).
(c) Do neither.
\item With some small probability, let individual entities mutate, i.e., switch their types to a value randomly sampled from $S$.
\item Repeat 4--7.
\end{enumerate}

In addition to the above steps, one may also include other factors such as spatial movement of entities and the carrying capacity of the environment, to make the dynamics more plausible and/or practical.

\subsection{Specific implementation}

We implemented a specific version of Hash Chemistry in Wolfram Research Mathematica. The following model settings were adopted:
\begin{itemize}
\item The possibility set of individual entities (types) are natural numbers ranging from 1 to 1,000, i.e., $S = \{1, 2, \ldots, 1000\}$.
\item The space is a two-dimensional continuous unit square with cut-off boundaries, i.e., entities are not allowed to move out of the boundaries.
\item The initial configuration is made of 10 individual entities of randomly chosen types, randomly distributed within the space.
\item Each simulation is run for 2,000 iterations (time steps).
\end{itemize}

\begin{figure}
\centering
\includegraphics[width=\columnwidth]{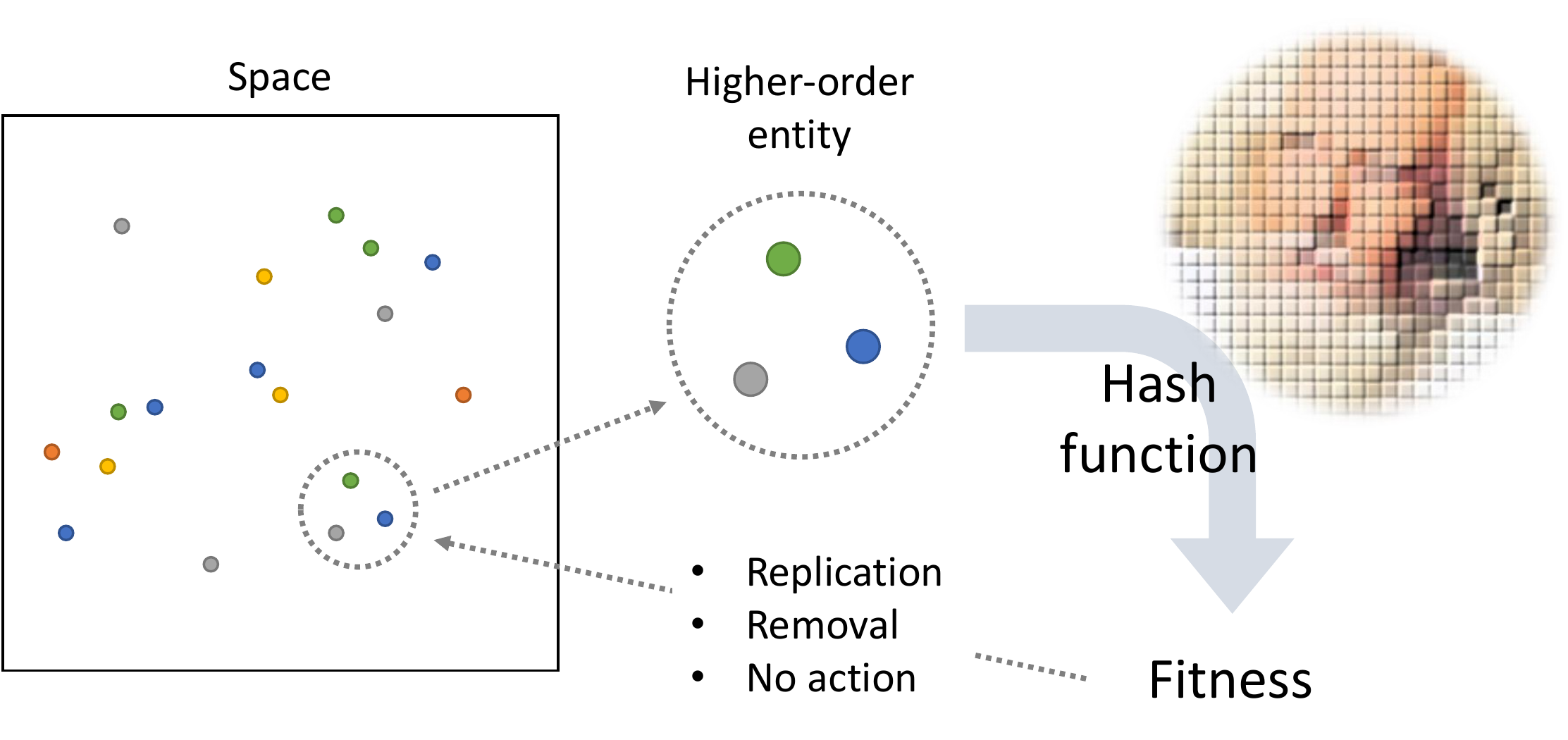}
\caption{Outline of the Hash Chemistry model. See text for details.}
\label{outline}
\end{figure}

In each iteration, the following steps are taken to update the system's state (also see Fig.~\ref{outline}):
\begin{enumerate}
\item Move each of the individual entities randomly by adding to its spatial position a small randomly oriented vector whose length is sampled from a half-normal distribution with $\sigma=0.15$.
\item For each of the positions of individual entities, do the following:
\begin{enumerate}
\item Create a set $N$ of individual entities that are of close distance (0.05 or less) from the focal position.
\item Choose a random subset $s$ of $N$ by randomly selecting $k$ entries from $N$, where $k$ is a random integer in $\{1, 2, \ldots, |N|\}$. This represents a multiset of entity types, i.e., a higher-order entity whose success in replication and/or survival is to be determined below.
\item With probability $1/|s|$ (this is to standardize the average probability of updating per entity per unit of time), do the following:
\begin{itemize}
\item Create a sorted list of types of the individual entities in $s$.
\item Calculate the fitness $f$ of $s$ by applying Mathematica's {\tt Hash} function \citep{hash}\footnote{In all the experiments presented in this paper, I used Mathematica version 11.3.0.0 on Windows (64-bit) and its {\tt Hash} command's ``Expression'' hash code type (which is the default choice of this function). As long as hash functions can generate a consistent hash value for entities of any size/order, the specific choice of hash code types would not alter the results fundamentally.} to the above list. The output is mapped to a $[0,1)$ fitness range by computing $(h \; \mathrm{mod} \; m) / m$, where $h$ is the output of {\tt Hash} ($m=100,000$ for the results shown here).
\item With probability $1-f$, delete all individual entities in $s$ from the space. {\em (death)}
\item With probability $f (1-|N|/d_\mathrm{max})$, where $d_\mathrm{max}$ is the maximum density of entities ($d_\mathrm{max} = 100$ for the results shown here), add copies of all individual entities in $s$ to the space. {\em (replication)}
\end{itemize}
\end{enumerate}
\item For each of the individual entities, change its type to a type randomly sampled from $S$ with probability 0.01. {\em (mutation)}
\item Randomize the order of individual entities.
\end{enumerate}

\section{Experiments}

\subsection{Main experiments}

\begin{figure}
\centering
\includegraphics[width=2.5in]{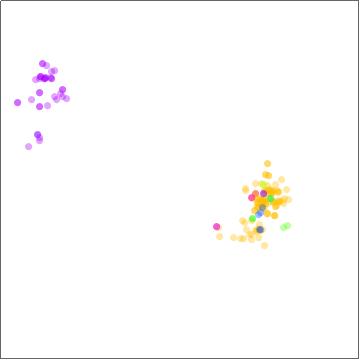}
\includegraphics[width=2.5in]{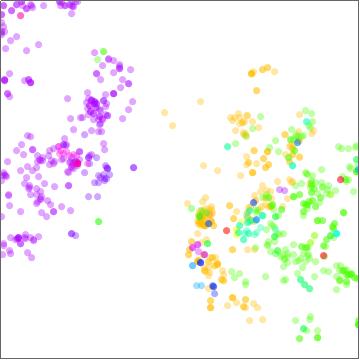}\\~\\
\includegraphics[width=2.5in]{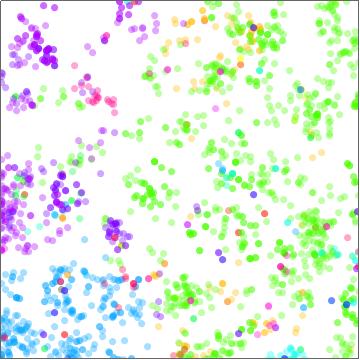}
\includegraphics[width=2.5in]{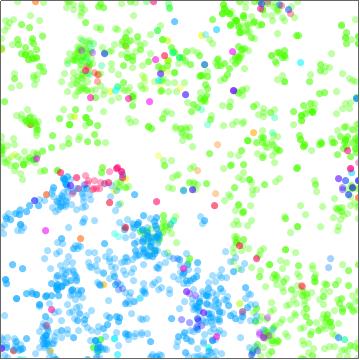}
\caption{Sample simulation run of Hash Chemistry. Each frame shows a snapshot of the system at a certain time point (from top left to bottom right: $t = 30, 100, 300, 1000$). Each individual entity is represented as a dot in the space, with a color showing its entity type. A movie of this simulation run is available on YouTube at \url{https://youtu.be/fVwUJ7pdPWY}.}
\label{fig:snapshots}
\end{figure}

We conducted Monte Carlo simulations of the Hash Chemistry model from randomly generated initial conditions. Sometimes the particle population became extinct in early stages of the simulation, and such results were excluded from analysis. We repeated conducting simulations until we obtained a total of 50 independent simulation runs that did not show particle extinction. As a result, we ended up running 65 independent simulations in total (i.e., probability of extinction: $(64-49)/64 = 23.4\%$). Figure \ref{fig:snapshots} shows a sequence of sample snapshots of the system's state and its spatio-temporal development over time taken from one illustrative simulation run (a movie of this simulation run is available on YouTube at \url{https://youtu.be/fVwUJ7pdPWY}). 

\begin{figure}
\centering
\includegraphics[width=4.5in]{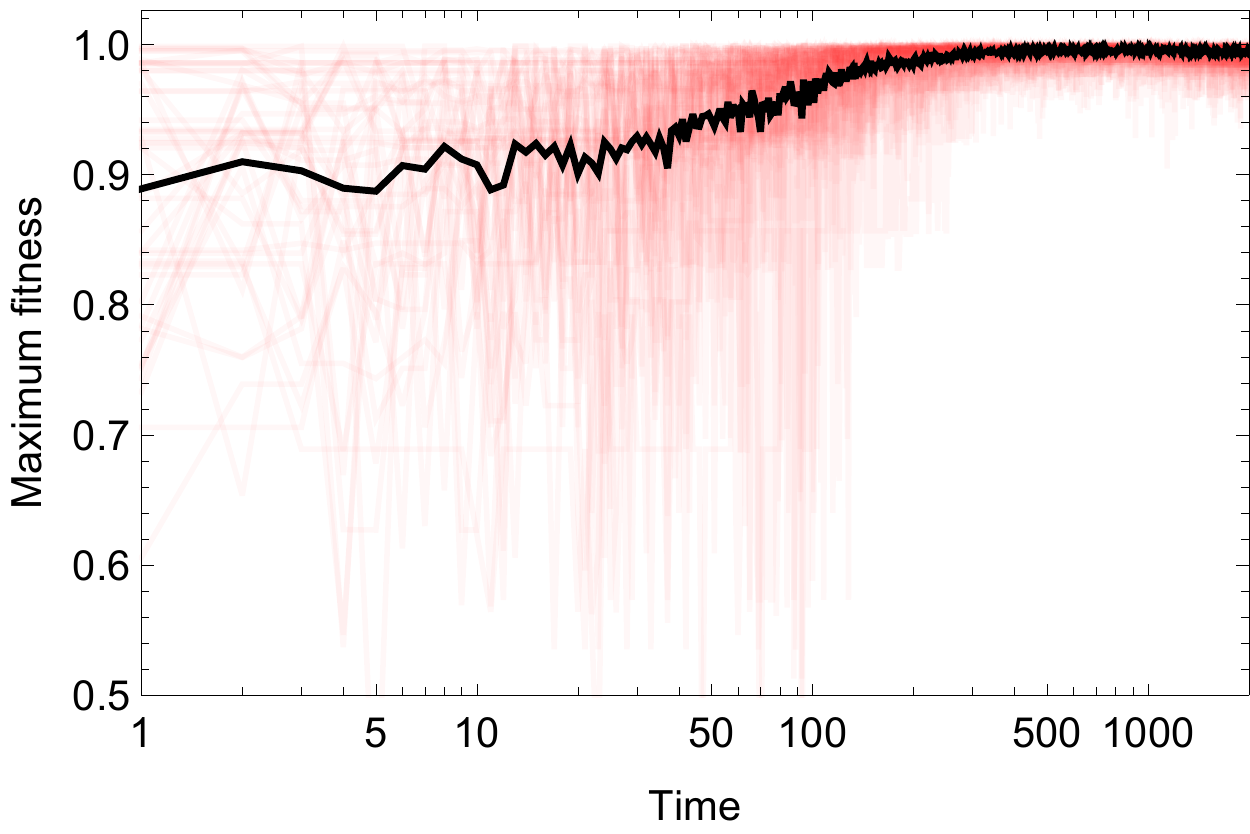}\\~\\
\includegraphics[width=4.5in]{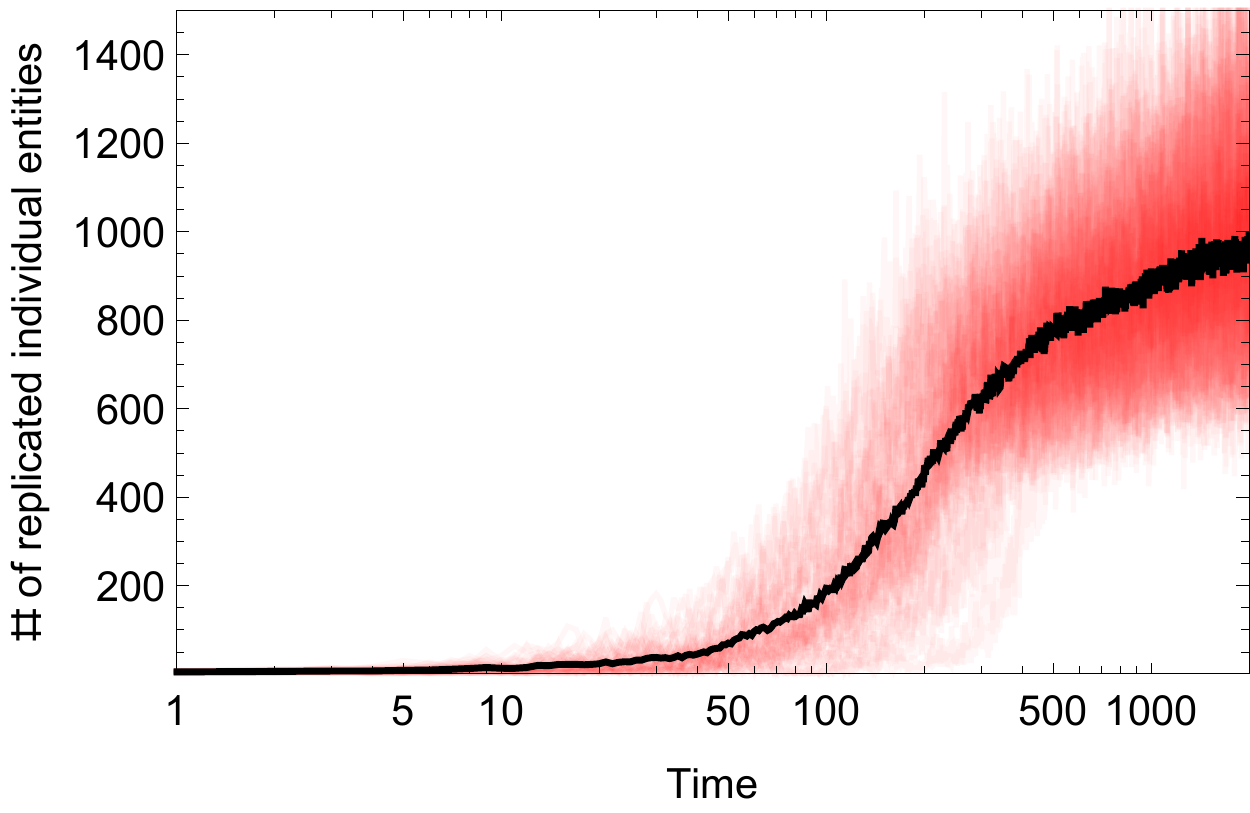}
\caption{Simulation results of Hash Chemistry seen in a conventional individual-based fitness view. Top: Maximum fitness value of replicated individual entities. Maximum fitness quickly increases and stays nearly at 1.0 (while the average fitness value remains around 0.7; results not shown). Bottom: Number of individual entities replicated in each time step. In each figure, the red curves show results of 50 independent simulation runs, while the black solid curve shows their average.}
\label{fig:adaptation}
\end{figure}

Figure \ref{fig:adaptation} presents time series of (i) maximum fitness values of individual entities that were successfully replicated, and (ii) the number of replicated individual entities. The former quickly increased and converged at the maximal value, while the latter gradually increased over time. These results show how the obtained dynamics would appear if seen through the lens of conventional individual-based fitness evaluations applied only to individual entities in $S$. The observed behaviors were neither surprising nor particularly different from what have been reported with many other evolutionary models (i.e., fitness simply went up over time).

\begin{figure}
\centering
\includegraphics[width=4.5in]{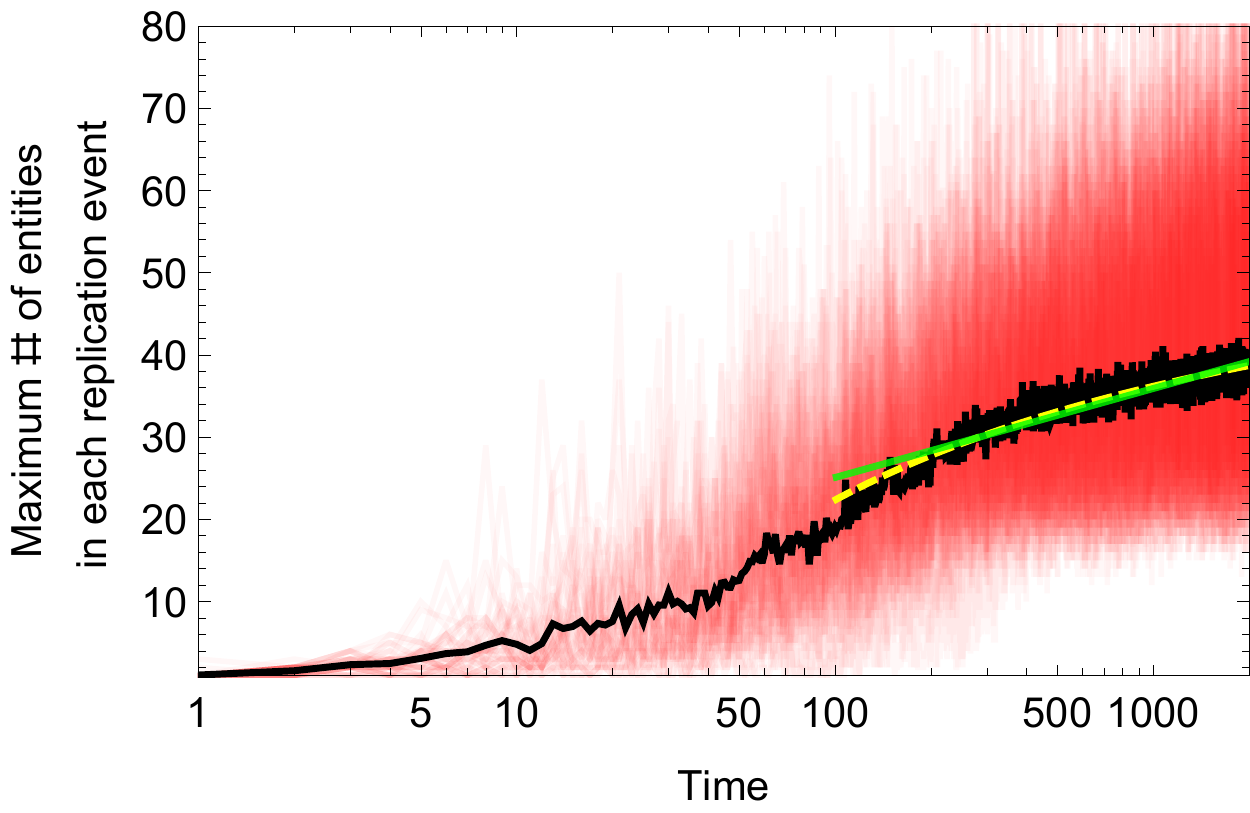}\\~\\
\includegraphics[width=4.5in]{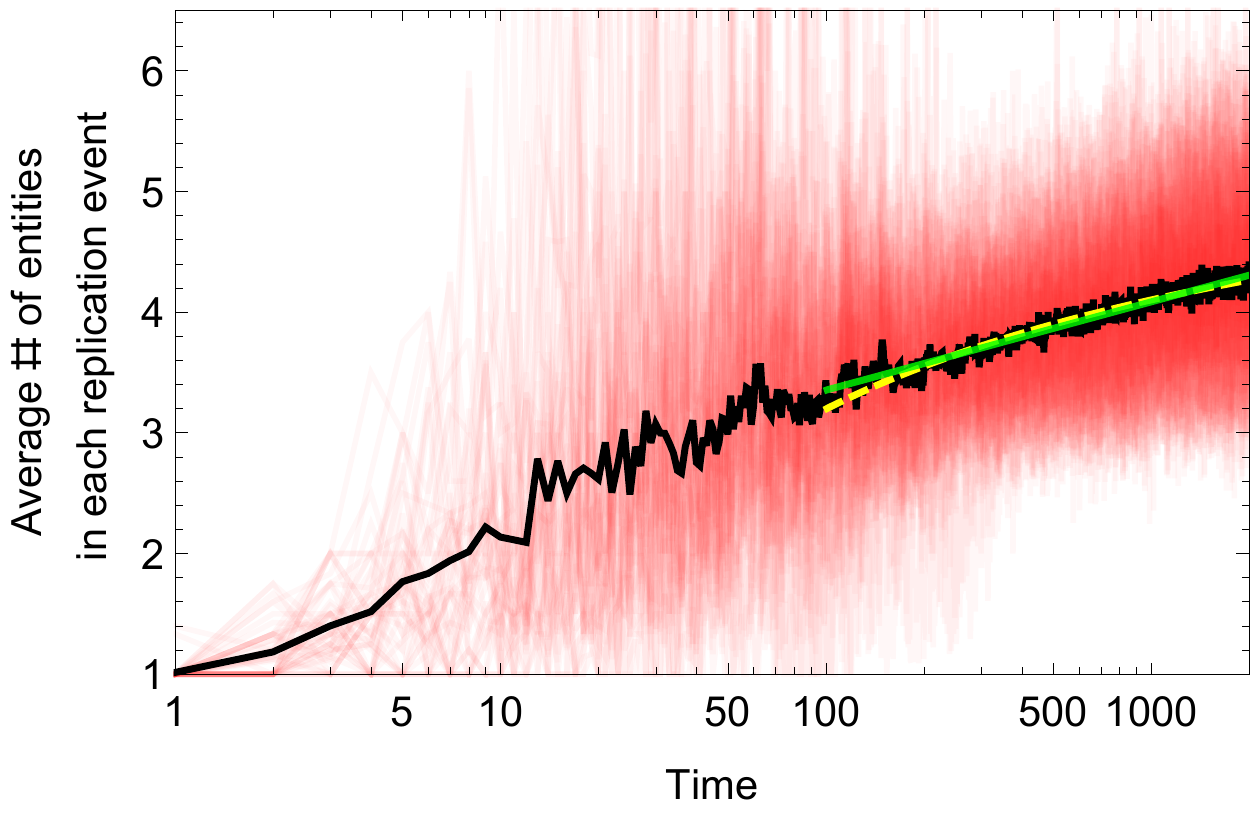}
\caption{Maximum (top) and average (bottom) numbers of individual entities involved in each replication event. In each figure, the red curves show results of 50 independent simulation runs, while the black solid curve shows their average. Yellow (dashed) and green (solid) curves are two different growth models (yellow: bounded growth, green: unbounded growth) fitted to the average behaviors during the time period 100--2,000. See Table \ref{tab:curvefits} for more details.}
\label{fig:higher-order}
\end{figure}

However, Figure \ref{fig:higher-order} tells a rather different story. In these plots, it is shown that the number of individual entities that were involved in a single replication event increased continuously over time\footnote{To characterize the trend of growth, two different growth models, bounded and unbounded ones, were fitted to the average behaviors during the time period 100--2,000 in logarithmic time scales. Details of the models and the results are shown in Table \ref{tab:curvefits}. The maximum number of individual entities (Fig.\ \ref{fig:higher-order} top) appeared more likely to be bounded probably because of the population density limit imposed in these simulations, while the average number (Fig.\ \ref{fig:higher-order} bottom) were more likely to be unbounded at least within the duration of these simulations.}. This means that the individual entities gradually became more and more replicated together with others, behaving as higher-order entities. This corresponds to {\em ongoing growth of complexity}, one of the behavioral hallmarks of OEE identified in \citep{taylor2016open}. This can be interpreted in that this evolutionary system continuously discovered slightly higher fitness values for increasingly higher-order entities in $S^*$ on the nontrivial fitness landscape defined by the hash function. Such spontaneous increase in the order of evolving entities is quite unique of this Hash Chemistry model.

\begin{table}
\centering
\caption{Summary of curve fitting of two different growth models to the results shown in Fig.\ \ref{fig:higher-order}}
\begin{tabular}{llll}
\hline
\multicolumn{1}{c}{Data} & Model & \multicolumn{2}{c}{Results} \\
\hline
Maximum number of & Bounded growth &  \multicolumn{2}{l}{Best fit:} \\
individual entities & $n(t) = -a / \log t + b$ & 
\multicolumn{2}{l}{$n(t) = -191.199 / \log t + 63.8593$} \\
involved in replication & & $R^2$ & $0.998259$\\
(Fig.\ \ref{fig:higher-order} top) & & AIC & $6882.23$ \\
 & & BIC & $6898.88$ \\
 \cline{2-4}
 & Unbounded growth &  \multicolumn{2}{l}{Best fit:} \\
 & $n(t) = a \log t + b$ & 
\multicolumn{2}{l}{$n(t) = 4.70574 \log t + 3.40787$} \\
 & & $R^2$ & $0.997697$\\
 & & AIC & $7413.38$ \\
 & & BIC & $7430.03$ \\
\hline
Average number of & Bounded growth &  \multicolumn{2}{l}{Best fit:} \\
individual entities & $n(t) = -a / \log t + b$ & 
\multicolumn{2}{l}{$n(t) = -12.494 / \log t + 5.90978$} \\
involved in replication & & $R^2$ & $0.999738$\\
(Fig.\ \ref{fig:higher-order} bottom) & & AIC & $-4963.37$ \\
 & & BIC & $-4946.72$ \\
 \cline{2-4}
 & Unbounded growth &  \multicolumn{2}{l}{Best fit:} \\
 & $n(t) = a \log t + b$ & 
\multicolumn{2}{l}{$n(t) = 0.317896 \log t + 1.88927$} \\
 & & $R^2$ & $0.999754$\\
 & & AIC & $-5079.77$ \\
 & & BIC & $-5063.12$ \\ 
\hline
\end{tabular}
\label{tab:curvefits}
\end{table}

\begin{figure}
\centering
\includegraphics[width=4.5in]{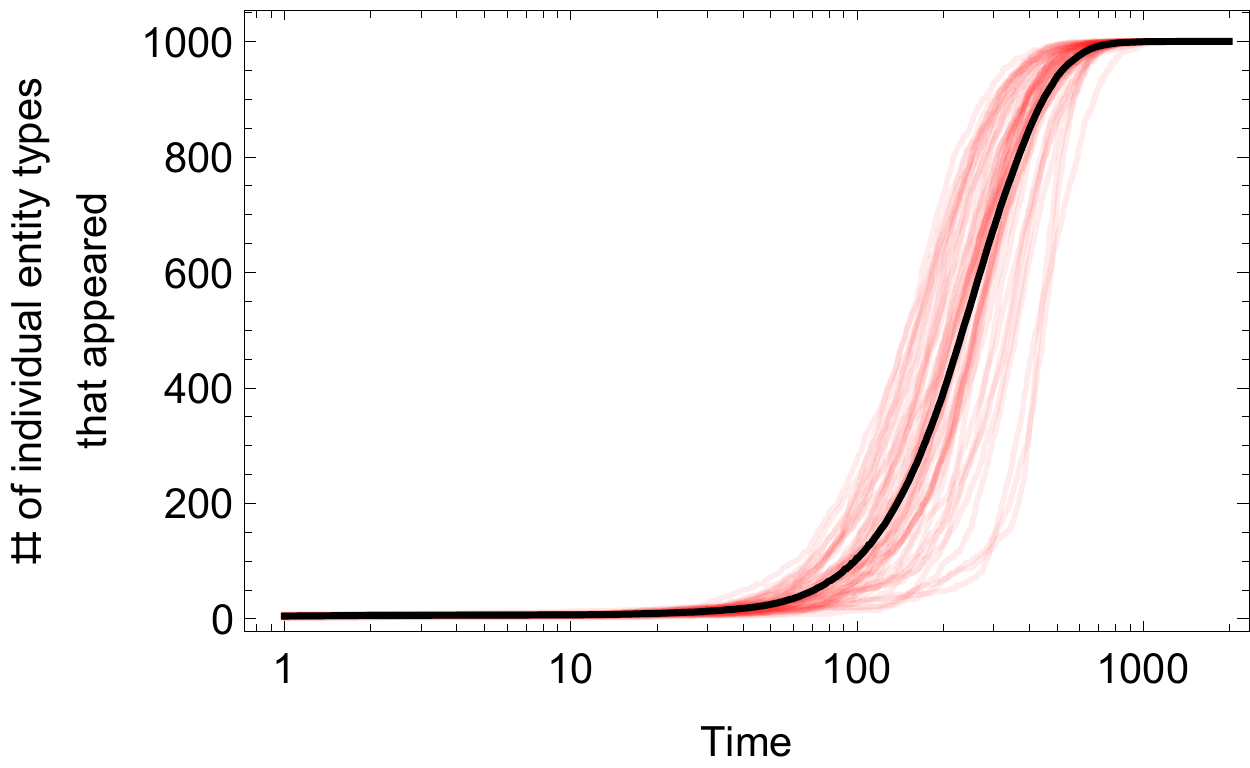}
\includegraphics[width=4.5in]{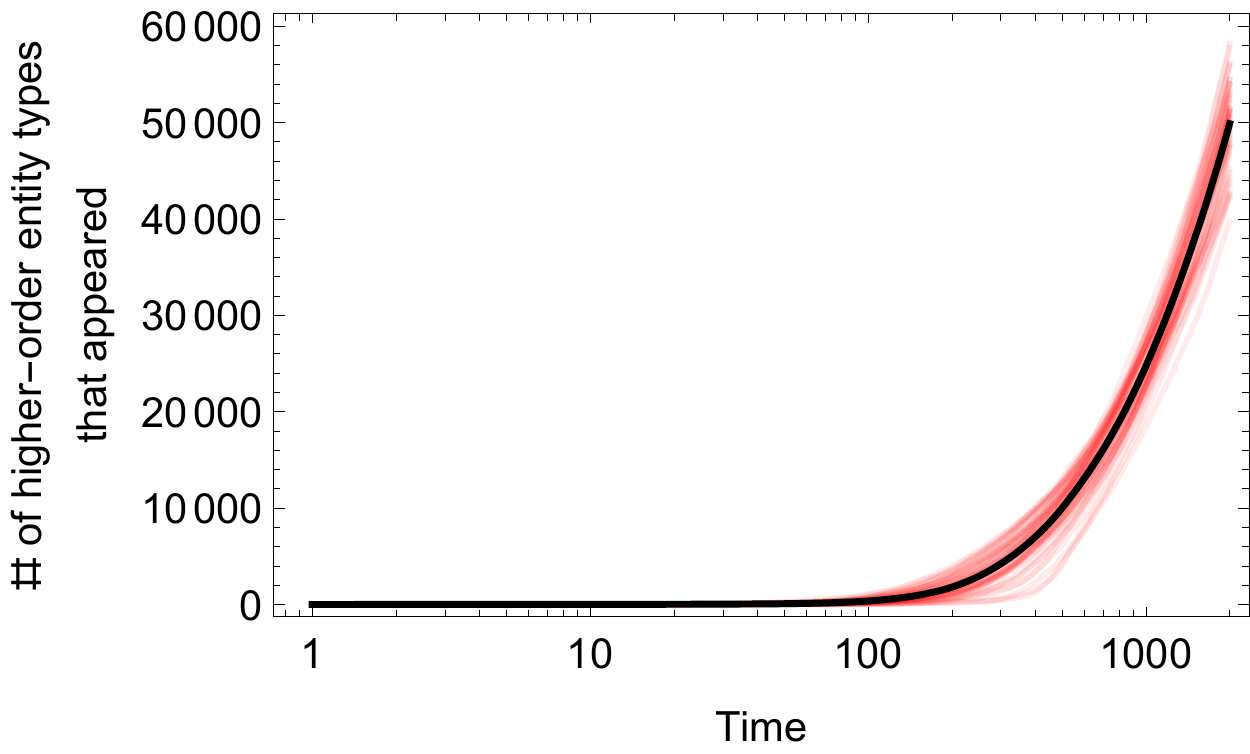}
\caption{Cumulative numbers of unique entity types that have ever appeared in the course of simulation. Top: Number of individual entity types, which saturates at the maximum value (1,000). Bottom: Number of higher-order entity types, which continues to grow almost linearly along time. In each figure, the red curves show results of 50 independent simulation runs, while the black solid curve shows their average.}
\label{fig:open-ended}
\end{figure}

The effect of the proposed cardinality leap can be visualized most directly by counting and plotting the cumulative numbers of unique entity types that have ever appeared in the course of the simulation run, as shown in Figure \ref{fig:open-ended}. If only the types of individual entities are counted (Fig.~\ref{fig:open-ended} top), the number grows following a typical logistic-like growth curve, quickly exhausting all the possible types (1,000) in the early stage of evolution. This corresponds to the inevitable consequence of evolutionary enumeration of possibilities. In this view, OEE is not possible. However, if the types of higher-order entities that successfully self-replicated are counted (Fig.~\ref{fig:open-ended} bottom), the number of possibilities continues increasing almost linearly along time, far above the number of possibilities of individual entities. This is made possible by the cardinality leap caused by formation of higher-order entities. Namely, the cardinality leap results in an unbounded number of combinations of individual entities, and such higher-order entities can and do appear as the system explores and discovers them as more fit entities in the course of evolution. 

\subsection{Control experiments}

Finally, to check if the observed behaviors reported above were truly nontrivial and relevant to OEE, control experiments were conducted by replacing the fitness evaluator with a random number generator that gives a newly generated random fitness value every time an entity is being evaluated. Unlike hash functions, such random number generators do not represent a consistent fitness landscape, which makes no room for adaptation to occur. Comparison of the results of these control experiments with the results obtained with Hash Chemistry should clarify which part of the results were truly driven by evolutionary adaptation.

\begin{figure}
\begin{tabular}{ll}
(a) & (b)\\
\includegraphics[width=0.48\columnwidth]{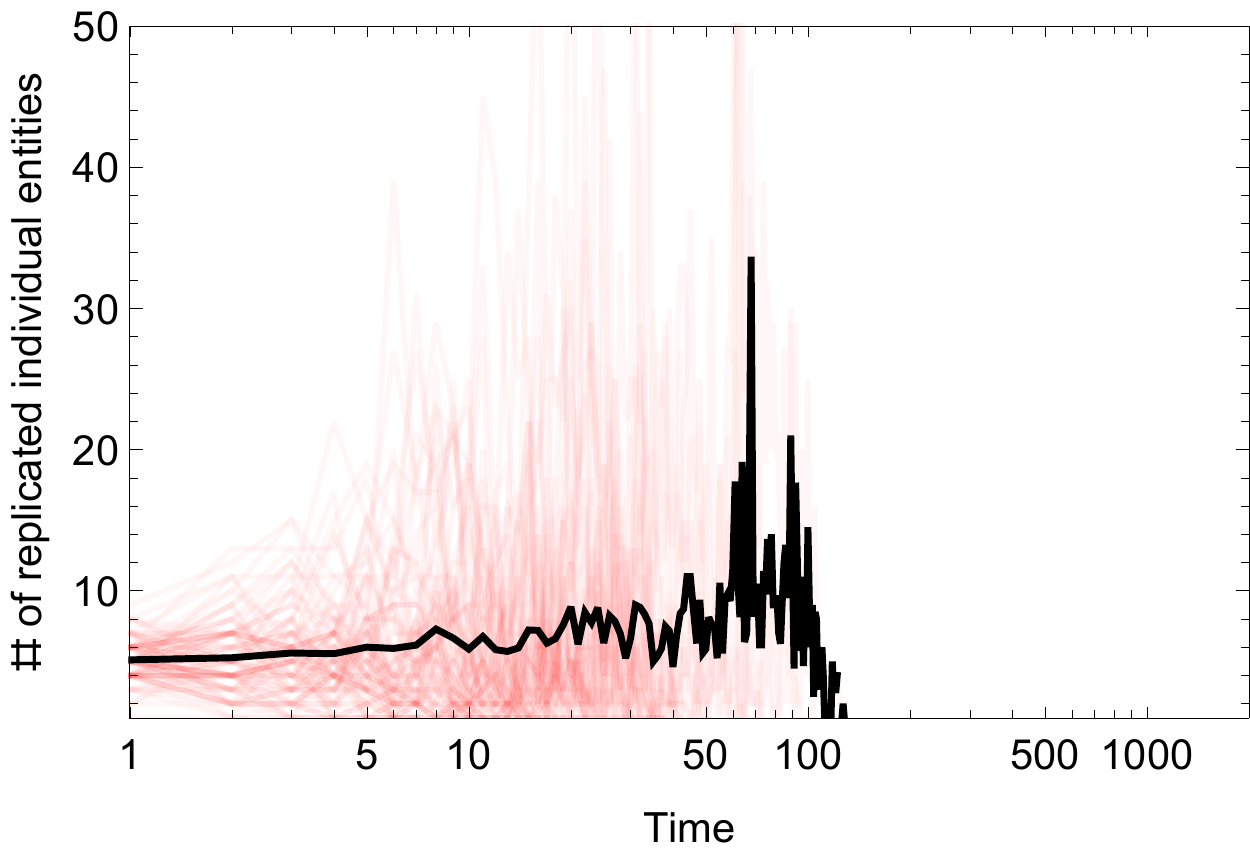}&
\includegraphics[width=0.48\columnwidth]{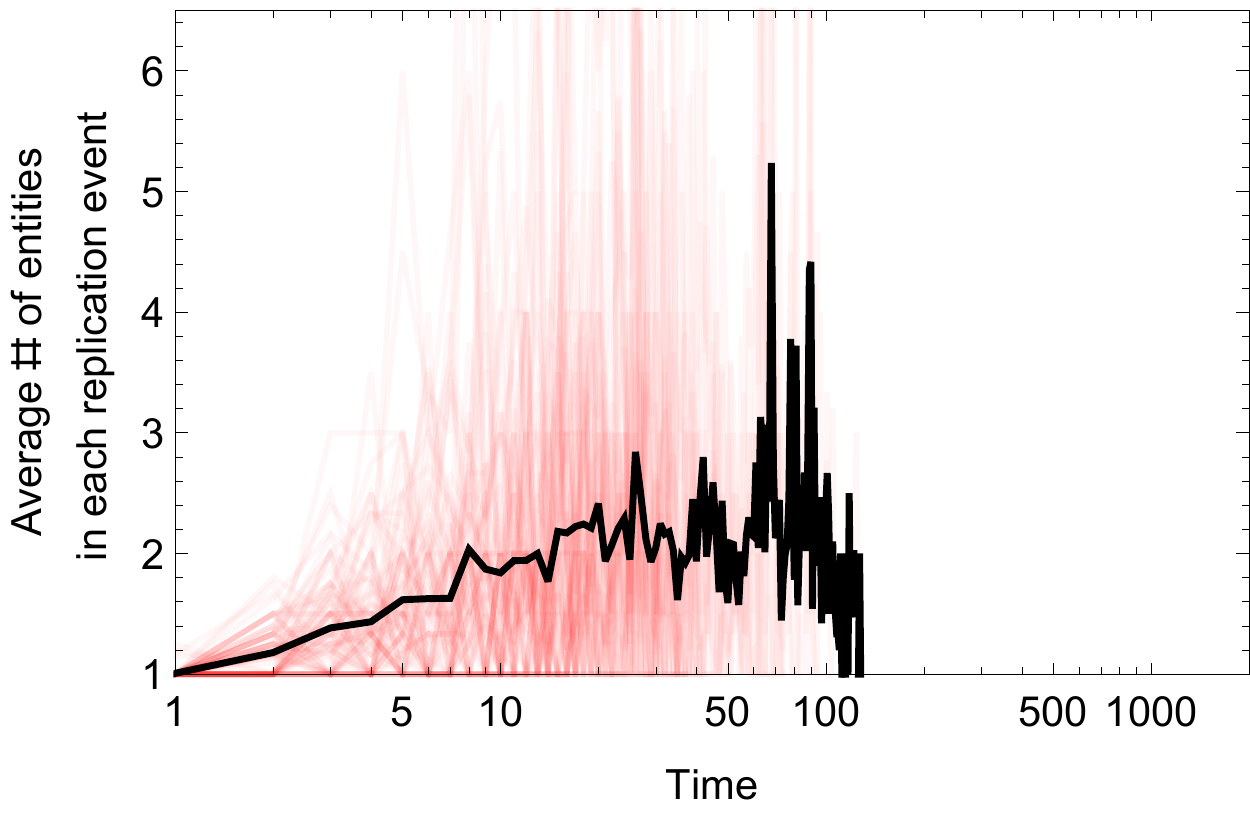}\\
 & \\
(c) & (d)\\
\includegraphics[width=0.48\columnwidth]{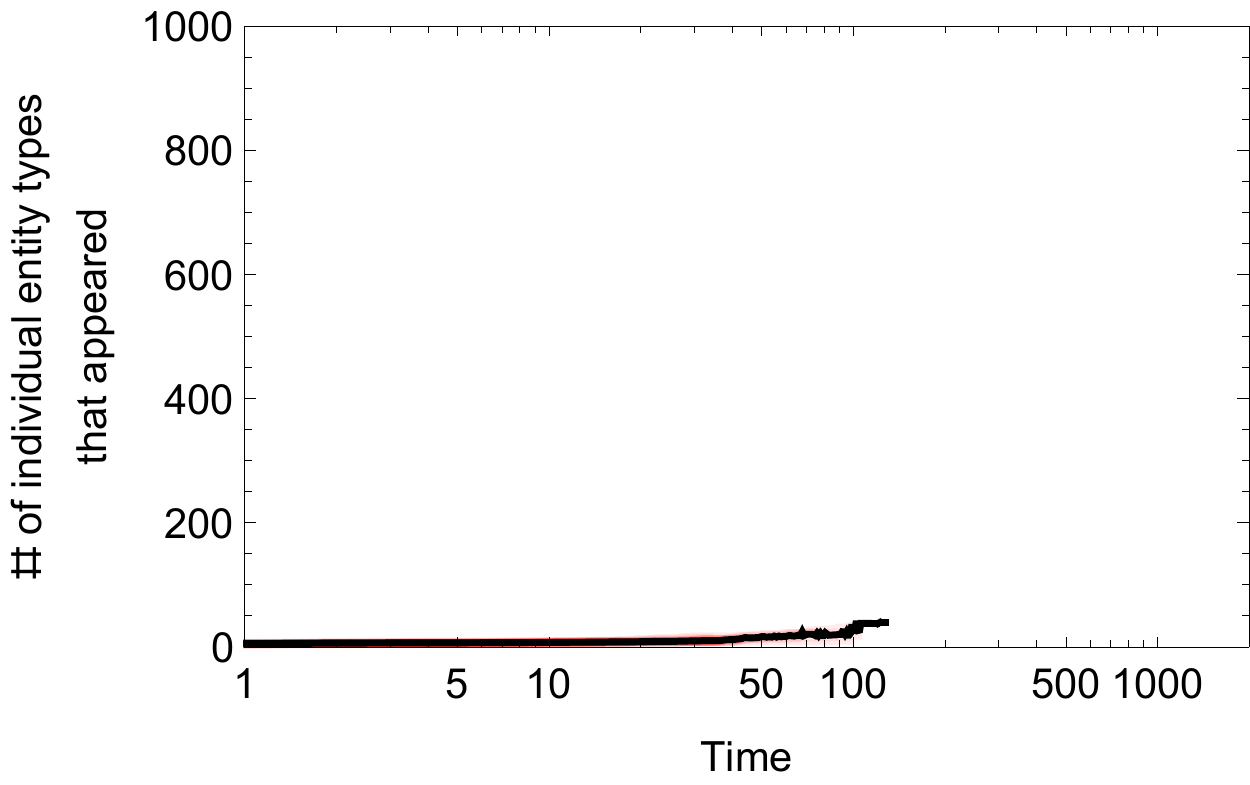}&
\includegraphics[width=0.48\columnwidth]{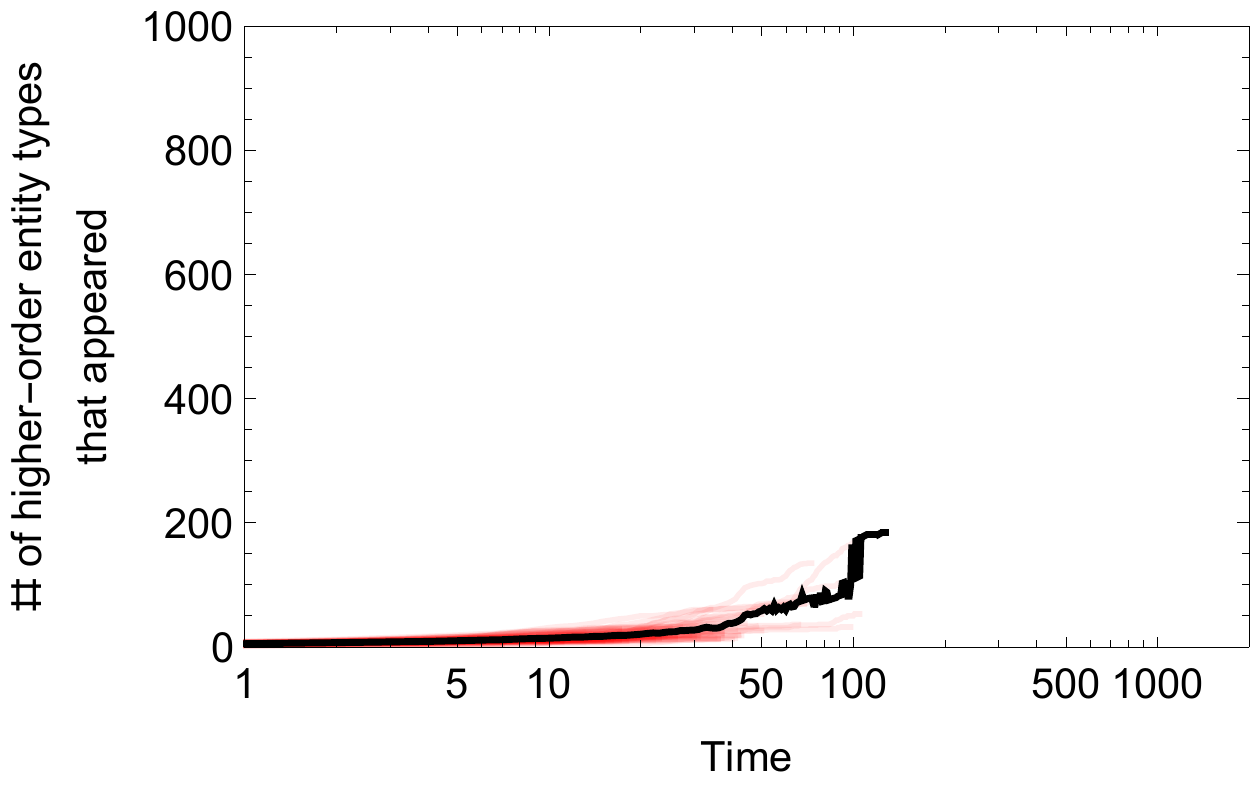}
\end{tabular}
\caption{Results of the first set of control experiments in which the fitness evaluator was replaced by a random number generator that returns a random fitness value sampled from a uniform distribution $[0,1]$. In each figure, the red curves show results of 100 independent simulation runs, while the black solid curve shows their average. In all the 100 runs, the population of entities became extinct in the early stages of simulation with no exceptions. (a) Number of individual entities replicated in each time step. (b) Average number of individual entities involved in each replication event. (c) Cumulative number of unique individual entity types. (d) Cumulative number of unique higher-order entity types.}
\label{fig:control}
\end{figure}

Figure \ref{fig:control} shows results from the first set of control experiments in which the fitness evaluator was replaced by a random number generator that returns a random fitness value sampled from a uniform distribution $[0,1]$. This is the fairest and most straightforward setting of control experiments since the distribution of hash values in the main experiments was also the same uniform distribution $[0,1]$. However, even though the average fitness value would be perfectly 0.5 in this setting (i.e., 50\% chance of successful replication), the population always collapsed into extinction within just the first few hundred steps of simulation, because of stochastic fluctuation and slightly negative selection pressure coming from the population density limit. The clear difference between this and the results of the main experiments strongly suggests that the sustained evolutionary dynamics observed in Hash Chemistry was indeed due to selection and adaptation.

The above control experiments always ended in population extinction, so the results did not provide much insight into whether or not the long-term behaviors observed in Hash Chemistry actually demonstrated {\em ongoing adaptive novelty,} another behavioral hallmark of OEE \citep{taylor2016open}. Therefore, another set of control experiments was conducted with a {\em biased} random number generator with an artificially inflated average fitness value to keep the population from extinction (and therefore, this was not quite fair as a control experiment). Specifically, random fitness values were sampled from a biased uniform distribution $[0.2,1]$, with the average fitness value being 0.6  (i.e., 60\% chance of successful replication). 

\begin{figure}
\begin{tabular}{ll}
(a) & (b)\\
\includegraphics[width=0.48\columnwidth]{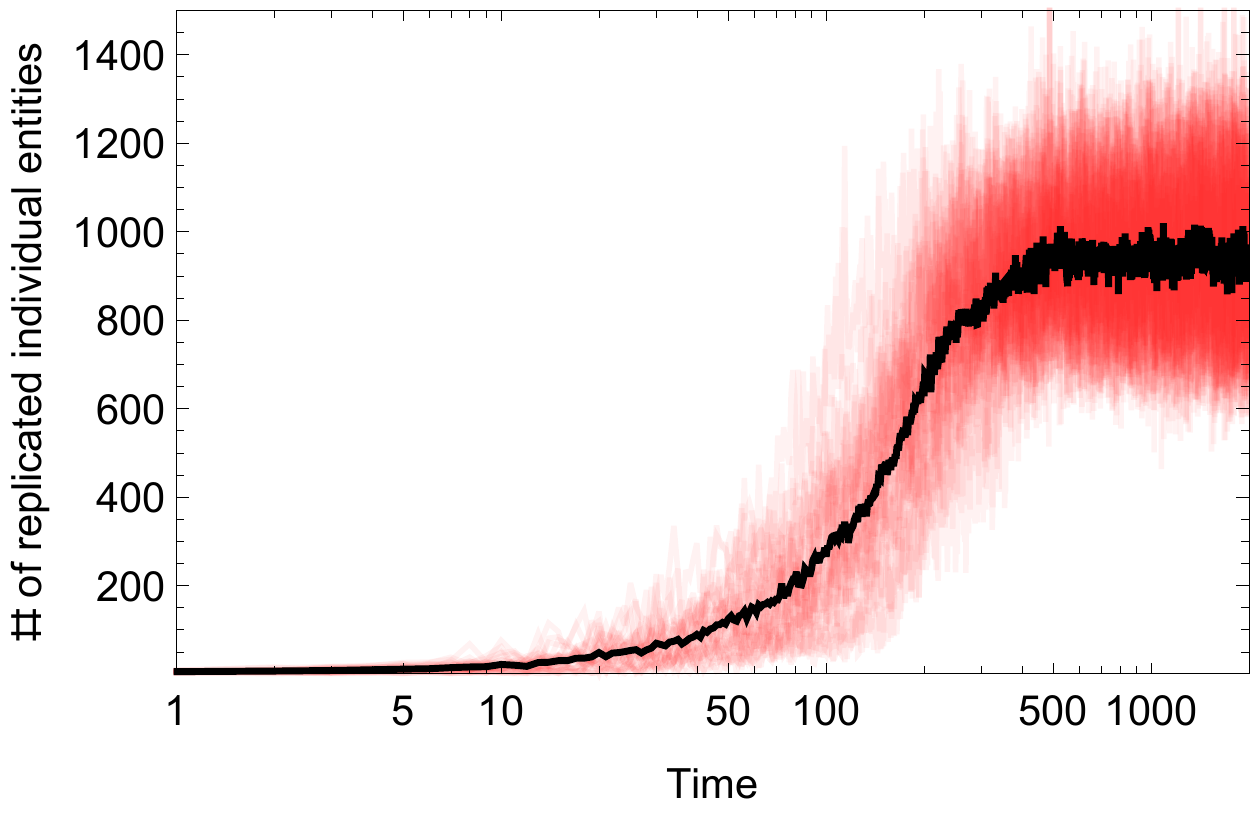}&
\includegraphics[width=0.48\columnwidth]{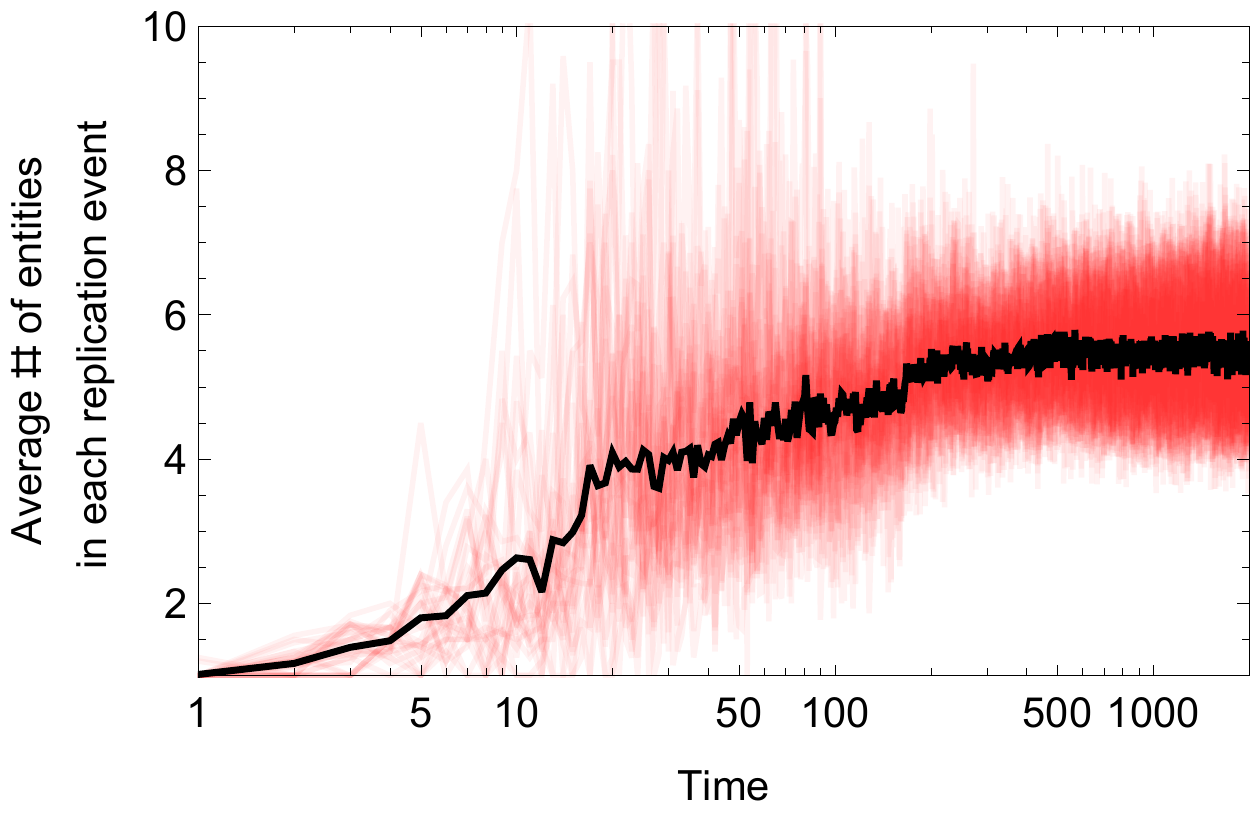}\\
 & \\
(c) & (d)\\
\includegraphics[width=0.48\columnwidth]{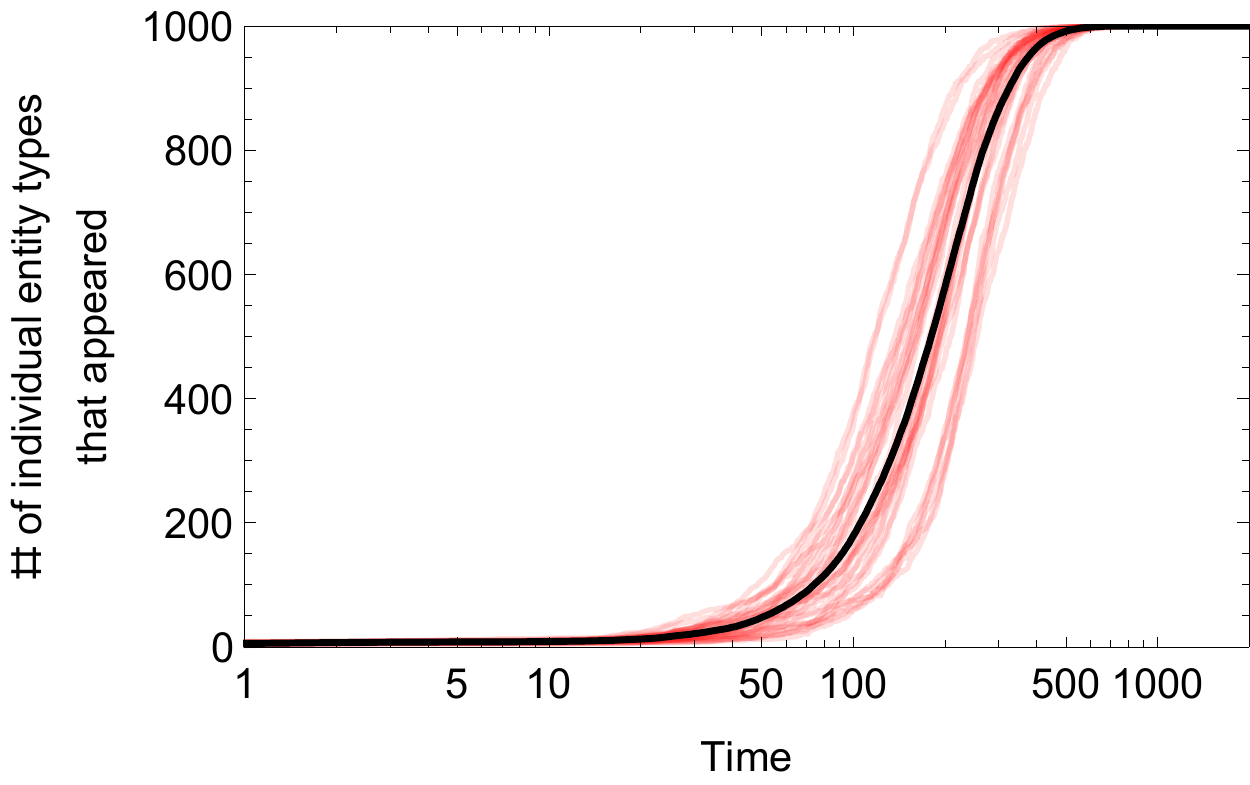}&
\includegraphics[width=0.48\columnwidth]{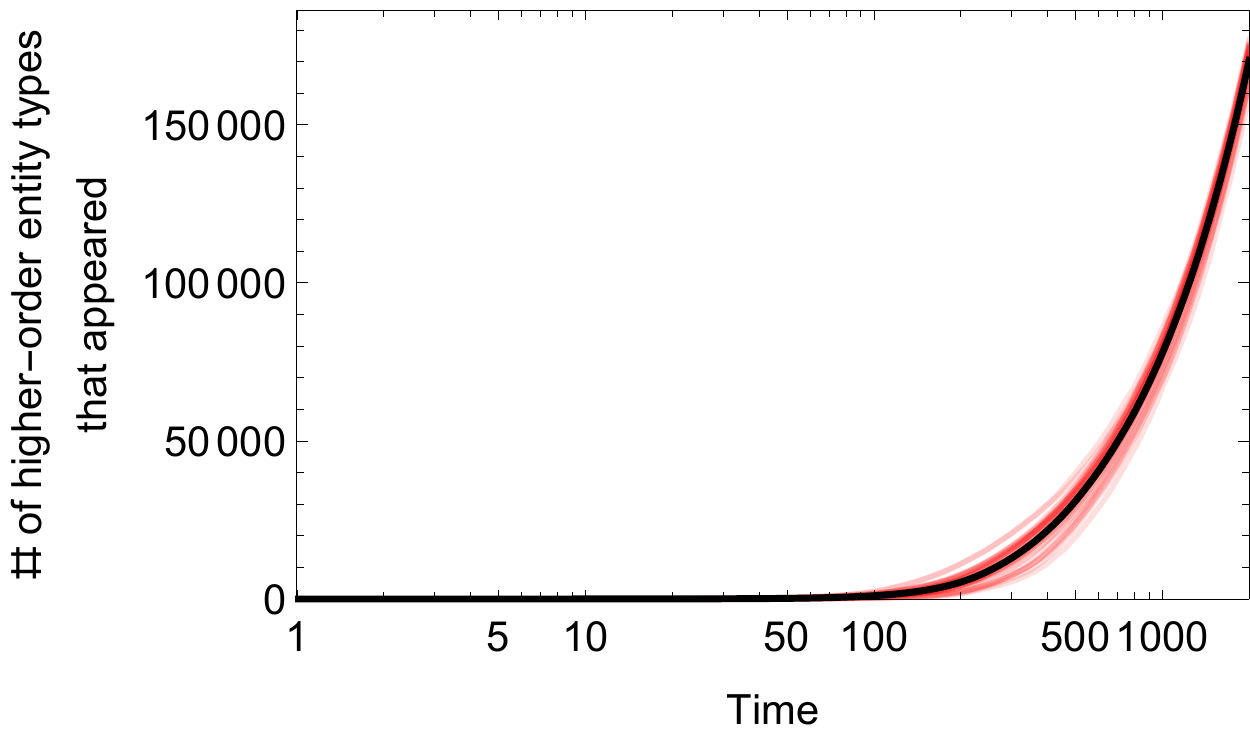}
\end{tabular}
\caption{Results of the second set of control experiments in which the fitness evaluator was replaced by a biased random number generator that returns a random fitness value sampled from a biased uniform distribution $[0.2,1]$. We repeated conducting simulations until we obtained a total of 30 independent simulation runs that did not show particle extinction. As a result, we ended up running 32 independent simulations in total (i.e., probability of extinction: $(31-29)/31 = 6.45\%$). In each figure, the red curves show results of 30 independent simulation runs, while the black solid curve shows their average. (a) Number of individual entities replicated in each time step. (b) Average number of individual entities involved in each replication event. (c) Cumulative number of unique individual entity types. (d) Cumulative number of unique higher-order entity types.}
\label{fig:control2}
\end{figure}

Figure \ref{fig:control2} shows the results of the second control experiments. It was seen that the number of novel higher-order entities did increase with time in this condition (Fig.\ \ref{fig:control2}(d)), indicating that this novelty production effect of the cardinality leap would not require selection or adaptation {\em per se.} However, the quantitative difference in this measure between the main and control experiments, i.e., that the numbers shown in Fig.\ \ref{fig:open-ended} bottom were only about one third of those in Fig.\ \ref{fig:control2}(d), indicates that there was strong selection going on in the main experiments with Hash Chemistry. Moreover, the average number of individual entities involved in single replication event converged to a fixed value and did not show continuous increase in the control experiments (Fig.\ \ref{fig:control2}(b)), in stark contrast to the main results obtained in Hash Chemistry (Fig.\ \ref{fig:higher-order} bottom). This strongly suggests that the continuous emergence of higher-order entities in Hash Chemistry was actual adaptation driven by selection.

\section{Conclusions}

In this paper, we emphasized the significance of formation of higher-order entities as a generalizable mechanism to induce a cardinality leap in the possibility set, naturally facilitating OEE. This idea was illustrated with a concrete computational model, Hash Chemistry, which adopted a general-purpose hash function as a means to evaluate fitness of evolving entities of any size or order. Numerical simulations successfully demonstrated evolutionary appearance of higher-order entities and unbounded increase in the cumulative number of novel types produced in evolution, even if the possibilities of individual entity types were finite. Comparison of these results with results of control experiments confirmed that the evolution of higher-order entities was driven by selection and adaptation. These results constitute a concrete, operationalized example of cardinality leaps through formation of higher-order entities, suggesting that it is indeed possible to achieve OEE in a relatively simple ALife model framework. 

Hash Chemistry as implemented in this paper is still quite simple and limited in many aspects, most importantly in terms of its large computational load. Because the population of particles can quickly grow in number, its numerical simulation would take a substantial amount of time (the entire set of the main experiments took one whole month to complete on a dedicated Windows workstation). The population density limit was introduced to keep the computational load at a manageable level, but it should be noted that this limit also altered the theoretical properties of the model. With this limit in place, the number of individual entities that could exist within an interaction range around a given location is essentially bounded, and therefore the possibility set of higher-order entities was still finite, strictly speaking. In this sense, Hash Chemistry was only an approximated demonstration of cardinality leaps.

Despite its technical simplicity and limitations, Hash Chemistry already has demonstrated some unique, non-trivial properties from an evolutionary systems viewpoint. Evaluating the fitness of higher-order entities at multiple scales means that the {\em context-dependent} fitness landscape is already present in this system. Specifically, how successful a particle (or an entity made of multiple particles) is in replication depends on what other particles/entities exist in its spatial vicinity, because they may be picked up together for fitness evaluation. Therefore, the effects of possible interactions between entities of any size/order are all represented implicitly in the hash function. Similarly, successful adaptation in this model requires adaptation at multiple scales, instead of fitness optimization at a particular scale of selection. Namely, even if a single particle has a high fitness value, it would not become evolutionarily dominant unless its combinations with multiple copies of itself as well as other types of particles also show high fitness values simultaneously, and such fitness criteria exist for combinations of entities of any size/order. This is an extremely complex, multiscale form of context dependence, which is beyond typical inclusive fitness or context-dependent fitness models that would make evolving entities adapt to other entities only at particular finite scales. In contrast, the adaptation seen in Hash Chemistry is the adaptation {\em at all scales at once,} which most of the computational evolutionary models have not demonstrated so far.

There are a number of directions of future research to be taken from the present study. First, a systematic theoretical study is needed to clarify which aspects of the Hash Chemistry model captures generalizable properties/mechanisms of OEE while which other aspects are specific only to this model. Second, concrete methodologies should be developed for designing and implementing more meaningful universal fitness evaluators (i.e., without using hash functions). Third, it is worth investigating how one can (and whether one should) make higher-order entities capable of more explicit self-maintenance, because the current model implementation of Hash Chemistry does not have an explicit representation of ``higher-order entities'' and thus they are not protected when a subset of particles are randomly selected. Fourth, it is not obvious whether one can/should seek even greater cardinality leaps by constructing a multiset of multisets of individual entities, a multiset of multisets of multisets, and so on. The effects of such hierarchical cardinality leaps on evolutionary dynamics are far from obvious. Finally, 
we should explore how the idea of cardinality leaps could be introduced to other existing ALife models to facilitate their open-endedness, including our recent evolutionary swarm models \citep{sayama2011seeking,sayama2018seeking}.

\section*{Acknowledgments}

The author thanks Howard H.\ Pattee for having insightful discussions that provided many inspirations at the initial stage of this work, and also two anonymous reviewers whose comments greatly helped improve the accuracy and clarity of this paper.

\bibliographystyle{apalike}
\bibliography{sayama-journal-rev}

\end{document}